\documentclass[11pt]{article}

\usepackage[utf8]{inputenc}
\usepackage[T1]{fontenc}
\usepackage{mathptmx}
\usepackage[margin=1in]{geometry}
\usepackage{amsmath,amssymb}
\usepackage{graphicx}
\usepackage{booktabs}
\usepackage{xcolor}
\usepackage{listings}
\usepackage{titlesec}
\usepackage{enumitem}
\usepackage{abstract}
\usepackage[hidelinks]{hyperref}
\usepackage{caption}
\usepackage{placeins}

\definecolor{linkblue}{RGB}{31,56,100}
\hypersetup{colorlinks=true, linkcolor=linkblue, citecolor=linkblue, urlcolor=linkblue}

\titleformat{\section}{\normalfont\large\bfseries\color{linkblue}}{\thesection}{0.6em}{}
\titleformat{\subsection}{\normalfont\normalsize\bfseries}{\thesubsection}{0.6em}{}
\titlespacing*{\section}{0pt}{1.4ex plus 1ex minus .2ex}{1ex plus .2ex}

\title{\vspace{-2em}\textbf{\color{linkblue}AuditWeave: A Tamper-Evident, Auditor-Navigable\\ Evidence Layer for AI-Assisted and Data-Transformation Workflows}}
\author{
  Vimal Nakrani\\
  \small Independent Researcher\\
  \small \href{https://pypi.org/project/auditweave/}{https://pypi.org/project/auditweave/}
}
\date{}

\begin{document}
\maketitle

\begin{abstract}
\noindent AI systems are increasingly used to assist consequential decisions in regulated domains such as auditing, finance, and healthcare. This creates a recurring obligation: an organization must be able to reconstruct, after the fact, which evidence informed a given conclusion, and to show that the record of that reasoning was not altered. Existing tools address related but distinct problems---model observability, drift monitoring, governance reporting---and are built for the machine-learning engineer operating a system, not the reviewer who must trace one specific conclusion back to its supporting evidence. We present \textbf{AuditWeave}, a lightweight Python library, with no runtime dependencies, that records the steps of AI-assisted and data-transformation workflows into a single append-only, hash-chained ledger. A small, system-agnostic event vocabulary spans both retrieval-augmented generation (RAG) pipelines and tabular/lakehouse transformations, so a conclusion that draws on both can be traced end-to-end through one record. Within a sealed ledger, any modification, reordering, insertion, or deletion of events is detectable through chain verification. We describe the design and evaluate recording overhead, scalability, and tamper-detection correctness on the reference implementation. The integrity guarantees cost tens of microseconds per event, and---as the hash-chain construction implies---verification flagged every injected mutation across four mutation classes over 2{,}000 randomized trials.
\end{abstract}

\section{Introduction}
Artificial-intelligence systems are moving from experimental tools to components embedded in consequential, regulated workflows. A growing share of this usage is \emph{assistive}: a model reads a body of source material and proposes a conclusion that a human then relies upon. Auditors use retrieval-augmented systems to navigate enormous transaction datasets; financial institutions apply AI to credit and fraud workflows; healthcare organizations use it to support prior-authorization decisions. In each case, the value of the system is inseparable from a question that regulators are increasingly asking: when an AI-influenced decision is made, what evidence informed it, and can the organization prove that the record of that reasoning is intact?

This question has two distinct parts. The first is \emph{provenance}: the ability to reconstruct, for any conclusion, the chain of source documents, retrievals, transformations, and model outputs that produced it. The second is \emph{integrity}: the ability to demonstrate that the recorded chain has not been altered---whether by accident or intent---after it was created. Both are prerequisites for the kind of accountability that regulated environments demand, and neither is fully addressed by current tooling.

Existing tools address related but distinct problems. Model-observability and monitoring platforms focus on operational concerns---latency, cost, drift, and performance---and are designed around the needs of the machine-learning engineer operating a system in production. Governance and model-risk platforms produce compliance reports and inventories. These are valuable, but they answer a different question than the reviewer's: \emph{given one conclusion, show me the evidence beneath it, in order, and assure me the record was not changed.} Source attribution---binding an AI output to the specific inputs that produced it---has been identified as a particularly weak link in current AI-audit implementations.

This paper presents \textbf{AuditWeave}, a small library, with no runtime dependencies, that addresses both parts directly. AuditWeave records workflow steps into an append-only, hash-chained ledger using a compact, system-agnostic event vocabulary. The same vocabulary describes a RAG pipeline and a tabular data-transformation job, so a single trail can span a workflow that mixes both---a property we argue is essential for realistic audit settings, where an AI-generated explanation often rests on upstream data aggregations. We make three contributions: (i) a lifecycle event model and reference architecture for cross-modal provenance; (ii) a tamper-evident ledger design with an associated verification procedure; and (iii) an empirical evaluation of overhead, scalability, and tamper-detection correctness on the open-source reference implementation.

\section{Background and Related Work}
Work adjacent to AuditWeave falls into four broad categories. \emph{Data-quality and validation frameworks} check datasets against declarative expectations and are mature and widely used, but they assess the state of data rather than recording the provenance of decisions made over it. \emph{Data-lineage and provenance systems} track how datasets and models are produced and transformed, providing valuable upstream context; recent work captures fine-grained, record- and attribute-level lineage in data-science and preparation pipelines~\cite{zhao2025lineage,belhajjame2025inmemory}, and provenance libraries for machine-learning systems serialize pipeline lineage in standardized forms such as W3C PROV~\cite{padovani2025provenance}. These systems are complementary to our aims, but they generally center on the data and model lifecycle rather than the reviewer's task of reconstructing a single conclusion, and they do not, in general, provide a unified tamper-evident record that links data transformations to model inferences and human sign-offs. \emph{Model-observability platforms} instrument production model behavior, emphasizing drift, performance, and cost, and target operators rather than reviewers.

Closest to our work are recent \emph{audit-trail and reasoning-auditing frameworks} for large language models. Ojewale et al.~\cite{ojewale2026audit} propose LLM audit trails as a sociotechnical mechanism for continuous accountability, contributing a lifecycle event framework, a reference architecture with lightweight emitters and append-only stores, and an open-source implementation; our event model and reviewer-oriented design are directly informed by, and intended to complement, this line of work, while extending the trail across both retrieval and tabular data-transformation steps. Other approaches pursue decentralized or blockchain-based auditing of model reasoning, providing consensus guarantees among multiple auditors at the cost of a distributed substrate~\cite{huang2025trust}, and data-provenance auditing techniques target dataset attribution for fine-tuned models~\cite{li2025dataprovenance}.

AuditWeave is distinguished by three deliberate choices. First, it is \emph{reviewer-oriented}: its primary operation is to take a conclusion and return the ordered evidence beneath it. Second, it is \emph{cross-modal}: one event vocabulary and one trail span RAG and tabular pipelines, so a conclusion that depends on both is traceable end-to-end. Third, it is \emph{integrity-first by construction}: every event is cryptographically chained to its predecessor, so tampering within a sealed ledger is detectable without reliance on an external service or distributed network~\cite{huang2025trust}. The library is deliberately small and carries no runtime dependencies---a stance that reflects a simple observation, that an accountability layer is adopted only when it is trivial to add to a pipeline that already exists.

Table~\ref{tab:compare} positions AuditWeave against representative systems from the surrounding categories. The comparison is qualitative; the systems differ in scope and were not built to the same end, so the point is not to rank them but to make the gap AuditWeave fills explicit. No single existing system combines tamper-evident storage, reviewer-oriented navigation, and a unified event model that covers both RAG and data-transformation provenance.

\begin{table}[htbp]
\centering
\small
\setlength{\tabcolsep}{4pt}
\renewcommand{\arraystretch}{1.15}
\begin{tabular}{lccccc}
\toprule
\textbf{Capability} & \textbf{AuditWeave} & \textbf{W3C PROV} & \textbf{MLflow} & \textbf{OpenLineage} & \textbf{LangSmith} \\
\midrule
Tamper-evident storage        & \checkmark & --- & --- & --- & --- \\
Reviewer-oriented navigation  & \checkmark & partial & --- & --- & partial \\
RAG provenance                & \checkmark & partial & --- & --- & \checkmark \\
Data-transformation provenance& \checkmark & \checkmark & partial & \checkmark & --- \\
Unified cross-modal event model & \checkmark & --- & --- & --- & --- \\
No runtime dependencies       & \checkmark & n/a & --- & --- & --- \\
\bottomrule
\end{tabular}
\caption{Qualitative positioning of AuditWeave against representative provenance, experiment-tracking, lineage, and LLM-observability systems. Entries reflect each system's \emph{primary design goals} rather than the absolute limit of what could be engineered with it: ``partial'' indicates a capability that is achievable with effort or covered only in part, and ``n/a'' marks cases where the row does not apply to a specification rather than a tool.}
\label{tab:compare}
\end{table}

\section{System Design}
\subsection{Definitions}
We use a few terms precisely throughout. An \emph{event} is an immutable record of a single step in a workflow, carrying a type, an actor, a payload, and links to the events it depends on. A \emph{trail} is an append-only, hash-chained sequence of events. \emph{Provenance}, for a given event, is the set of upstream events reachable by following its links---its evidentiary ancestry. \emph{Integrity} is the property that a sealed trail has not been altered since creation, and is established by \emph{verification}, the recomputation of the hash chain. An \emph{evidence record} is the view a reviewer obtains for a particular conclusion: that conclusion together with its provenance and an integrity result. These definitions are deliberately operational---each corresponds directly to a construct in the implementation---rather than formal in a logical sense.

\subsection{Event model}
The core abstraction is the \emph{event}: an immutable record of one step in a workflow. AuditWeave defines six event types, chosen to be broad enough to describe both an LLM retrieval pipeline and a data-transformation job using a single vocabulary:
\begin{itemize}[leftmargin=1.4em,itemsep=2pt,topsep=3pt]
  \item \textbf{Source} --- a raw input document or dataset.
  \item \textbf{Retrieval} --- the selection of source material to answer a query.
  \item \textbf{Transformation} --- a data transformation step.
  \item \textbf{Inference} --- a model producing an output.
  \item \textbf{Decision} --- a conclusion or classification.
  \item \textbf{Attestation} --- a human reviewing and signing off on a prior event.
\end{itemize}
Each event records the actor responsible (a human, a model, or a system component), a payload of type-specific fields, optional labels, and---critically---a set of \emph{links} to the upstream events on which it depends. The links form a directed acyclic graph that captures provenance: a decision links the inference that produced it; that inference links the retrieval that supplied its context; that retrieval links the source documents it selected. Walking these links backwards from any event yields its complete evidentiary ancestry.

To support confidential data, an event may bind to the SHA-256 hash of a source document rather than its contents. This allows the trail to prove which exact bytes informed a decision without storing the document itself---important in regulated settings where the underlying data cannot be copied into an auxiliary store.

\subsection{The tamper-evident ledger}
Events are sealed into a \emph{Trail}, an append-only sequence linked by a SHA-256 hash chain. Each event stores the hash of the previous event together with its own hash, which is computed over its content including the predecessor hash and its sequence position. Because every hash incorporates the prior hash, any modification, insertion, deletion, or reordering of events invalidates the hashes of all subsequent events. The construction is conceptually similar to distributed ledgers and version-control histories, applied here to a narrow and practical problem: demonstrating that an evidence record was not altered after creation. It needs no distributed network. A single append-only file and the verification routine are enough for a reviewer to detect tampering.

Verification recomputes the entire chain and reports, for every event, whether its stored hash matches a fresh recomputation of its content and whether its recorded predecessor hash matches the actual previous event. A failure of either check localizes the tampering to a specific event. The following illustrates the recording and verification interface:

\begin{lstlisting}[language=Python]
from auditweave import Recorder
rec = Recorder()
s = rec.source("ledger.csv", content_hash=...)
t = rec.transformation("aggregate by quarter", [s.id])
d = rec.decision("Q3 revenue = $4.2M", [t.id])
assert rec.trail.verify().ok    # chain intact
\end{lstlisting}

\subsection{Navigation}
Where the ledger stores evidence, the \emph{navigator} presents it in the form a reviewer requires: given a conclusion, it returns the full upstream ancestry in reading order, each event summarized in a single line, accompanied by an integrity stamp indicating whether the record verifies. This is the operation an auditor performs when asked to justify a conclusion---reconstructing, from one statement, the complete trail of evidence beneath it, with assurance that the trail has not been altered.

\subsection{Adapters}
Two adapters translate framework-specific objects into the common event vocabulary. The RAG adapter records a full retrieve--prompt--generate--conclude turn, binding each retrieved chunk to its content hash. The tabular adapter records dataset registration and transformation lineage and accepts lightweight descriptors so it works with Spark, pandas, dbt models, or SQL jobs without requiring any of them as a dependency. Because both adapters write into the same trail, a conclusion that depends on both an AI inference and upstream data transformations is traceable end-to-end through one record.

\subsection{Threat model}
It is worth being explicit about what AuditWeave assumes and what it does not attempt, since the value of an integrity claim rests entirely on its scope.

\noindent\textit{Assumptions.} Events are recorded honestly at capture time---the trail attests to what was recorded, not to whether the recording faithfully reflected the world. Once an event is sealed into the chain it is treated as immutable. An adversary may later attempt to modify, reorder, insert, or delete events in the stored ledger.

\noindent\textit{Guarantee.} Under these assumptions, any such alteration within a sealed ledger is surfaced by chain verification, which localizes the inconsistency to a specific event.

\noindent\textit{Non-goals.} AuditWeave does not prevent tampering; it makes tampering evident. It does not verify the correctness or truthfulness of captured information. And it does not, on its own, defend against an adversary who controls the storage medium and recomputes the entire chain after a modification---closing that gap requires anchoring the chain head to an external, append-only reference, which we discuss in Section~\ref{sec:discussion} and leave to future work. Distributed or multi-party settings, where consensus among auditors is desired, are out of scope; that design point is occupied by other systems~\cite{huang2025trust}.

\section{Evaluation}
We evaluate three properties of the reference implementation that are objectively measurable and relevant to deployment: recording overhead, scalability, and tamper-detection correctness. All measurements were produced by an automated benchmark suite distributed with the library, executed against the published version of the package. Timings use a high-resolution monotonic clock and report best-of-repeats to reduce noise.

Measurements were taken on an x86-64 Linux environment (Intel Xeon-class CPU at 2.8\,GHz, Python 3.12) running in a containerized single-core configuration. Absolute timings should therefore be read as indicative rather than as hardware-optimized figures; the quantities of interest---how cost scales with trail size, and whether tampering is detected---are properties of the algorithm and are reproducible across environments via the published suite. We report per-event costs and asymptotic behavior rather than peak throughput for this reason.

\subsection{Recording overhead}
We compare full hash-chained recording against a no-integrity baseline---appending plain dictionaries to a Python list---to isolate the marginal cost of the integrity guarantees. Over 10{,}000 events, AuditWeave records at \textbf{22.42 microseconds per event}, against \textbf{0.24 microseconds} for the baseline. The relative increase is large ($93\times$), but the absolute cost is what matters in practice: at roughly 44{,}600 events per second on a single core, the integrity overhead is unlikely to be a practical bottleneck in realistic AI-assisted workflows, where individual steps---model inference, retrieval, I/O---typically run from tens to hundreds of milliseconds.

\subsection{Scalability}
\begin{figure}[htbp]
\centering
\includegraphics[width=0.66\textwidth]{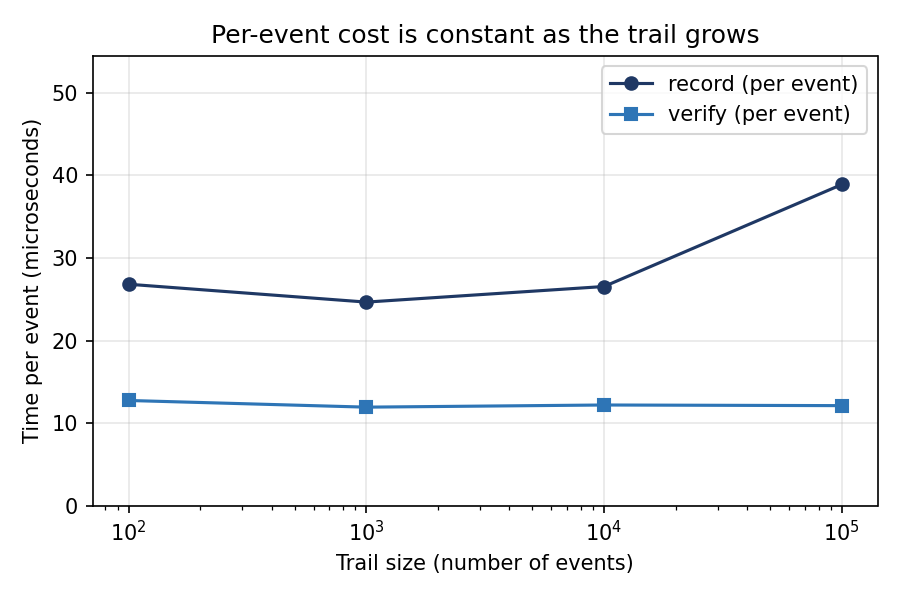}
\caption{Per-event cost of recording and verification remains constant as the trail grows across three orders of magnitude.}
\label{fig:scale}
\end{figure}

\begin{figure}[htbp]
\centering
\includegraphics[width=0.66\textwidth]{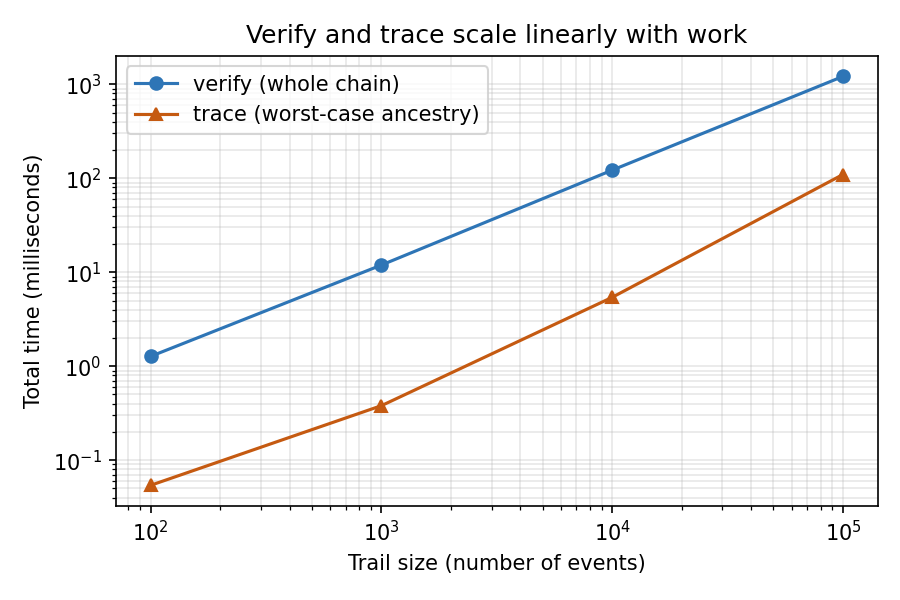}
\caption{Total verification and worst-case tracing time scale linearly with the amount of work performed.}
\label{fig:verify}
\end{figure}

We measure how recording, verification, and worst-case tracing scale as the trail grows from 100 to 100{,}000 events (Table~\ref{tab:scale}, Figures~\ref{fig:scale}--\ref{fig:verify}). Recording cost per event remains effectively constant; verification is linear in the number of events, as expected for a full chain recomputation; and worst-case tracing---retrieving the complete ancestry of the final event in a fully linear chain---grows linearly with the size of that ancestry.

\begin{table}[htbp]
\centering
\small
\begin{tabular}{rrrrr}
\toprule
\textbf{Events} & \textbf{Record ($\mu$s/ev)} & \textbf{Verify ($\mu$s/ev)} & \textbf{Verify total (ms)} & \textbf{Trace worst (ms)} \\
\midrule
100 & 26.82 & 12.74 & 1.27 & 0.05 \\
1{,}000 & 24.66 & 11.93 & 11.93 & 0.38 \\
10{,}000 & 26.54 & 12.19 & 121.95 & 5.41 \\
100{,}000 & 38.90 & 12.12 & 1211.69 & 109.60 \\
\bottomrule
\end{tabular}
\caption{Scalability of recording, verification, and worst-case tracing across trail sizes.}
\label{tab:scale}
\end{table}

Even at 100{,}000 events, full-chain verification completes in \textbf{1.21 seconds}, confirming that integrity checking remains practical at scale. Worst-case tracing of the entire ancestry completes in \textbf{109.6 milliseconds}; in practice, the ancestry of a typical conclusion is far smaller than the whole trail, so real navigation queries are substantially faster.

\subsection{Tamper-detection correctness}
\begin{figure}[htbp]
\centering
\includegraphics[width=0.66\textwidth]{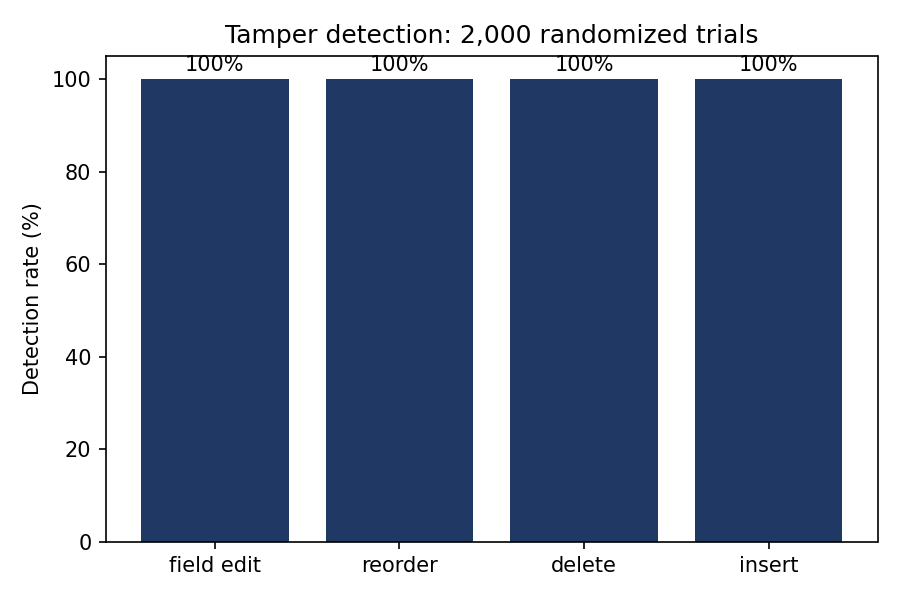}
\caption{Detection rate by mutation class. Verification flagged every injected mutation in all four classes, as the chain construction requires.}
\label{fig:tamper}
\end{figure}

The central guarantee of AuditWeave is that tampering with a sealed trail is detectable. This follows from the hash-chain construction, so the purpose of the experiment is validation---confirming the implementation behaves as the design requires---rather than the discovery of a surprising result. We test four classes of mutation: editing a field in an event, reordering two events, deleting an event, and inserting a forged event. Each class is run over 500 randomized trials, for 2{,}000 in total. In every trial a mutation is applied to a freshly built trail and verification is run; detection is recorded only if verification reports failure (Table~\ref{tab:tamper}, Figure~\ref{fig:tamper}).

\begin{table}[htbp]
\centering
\small
\begin{tabular}{lrrr}
\toprule
\textbf{Mutation class} & \textbf{Trials} & \textbf{Detected} & \textbf{Rate} \\
\midrule
field edit & 500 & 500 & 100\% \\
reorder    & 500 & 500 & 100\% \\
delete     & 500 & 500 & 100\% \\
insert     & 500 & 500 & 100\% \\
\bottomrule
\end{tabular}
\caption{Tamper-detection results across four mutation classes, 500 trials each.}
\label{tab:tamper}
\end{table}

As expected from the hash-chain construction, verification detected every injected mutation across all 2{,}000 trials. The mechanism is worth spelling out, since each class fails the chain differently. A field edit changes the content hash of the mutated event. A reorder or a deletion breaks the predecessor-hash relationship and shifts sequence positions. An inserted forged event is the subtlest case: even when its own predecessor hash is set correctly, it displaces the genuine following event, whose recorded predecessor hash no longer matches its new neighbor, and the downstream sequence numbers shift by one. In each case the construction forces the alteration to surface during verification rather than relying on it being noticed.

\section{Discussion and Limitations}\label{sec:discussion}
AuditWeave provides tamper-\emph{evidence}, not tamper-\emph{resistance}: it guarantees that alteration of a sealed record is detectable, not that alteration is prevented. An actor who controls the storage medium could in principle recompute the entire chain after a modification; defending against this requires anchoring the chain head to an external, append-only reference---for example, periodic publication of the head hash to an independent timestamping service---which we identify as future work. The current design also assumes honest recording at capture time: it attests to what was recorded, not to whether the recording faithfully reflected reality. This is the appropriate scope for an evidence layer, which documents a process rather than adjudicating its correctness.

The library does not, by itself, establish compliance with any particular regulation; whether a given trail satisfies a specific regulatory requirement is a determination for auditors and counsel. Its contribution is infrastructural: it supplies the verifiable, navigable substrate on which such determinations can be made. Planned extensions include automatic instrumentation adapters for common LLM frameworks, per-reviewer signed attestations using public-key cryptography, exportable human-readable evidence reports, and Merkle-root anchoring for external timestamping.

\section{Conclusion}
We presented AuditWeave, a lightweight, self-contained evidence layer that records AI-assisted and data-transformation workflows into a single tamper-evident, auditor-navigable ledger. By unifying retrieval and data-transformation provenance under one event vocabulary and chaining every event cryptographically, AuditWeave allows any conclusion to be traced to its supporting evidence with assurance that the record is intact. Our evaluation shows that these guarantees impose a per-event cost of tens of microseconds, scale to hundreds of thousands of events, and surface every injected tampering attempt across four mutation classes. The reference implementation is available as open-source software, and we believe it provides a practical foundation for accountability in the growing class of AI-assisted decision workflows in regulated domains.

\section*{Availability}
AuditWeave is released as open-source software under the Apache~2.0 license and is installable from the Python Package Index (\href{https://pypi.org/project/auditweave/}{pypi.org/project/auditweave}). The benchmark suite used to produce all results in this paper is distributed alongside the library and is fully reproducible via a single command.

\FloatBarrier

\end{document}